\documentclass[letterpaper,twocolumn]{article} 

\usepackage[top=1in, bottom=1.25in, left=1.25in, right=1.25in]{geometry}

\usepackage[pdftex]{graphicx}
\DeclareGraphicsExtensions{.pdf}
\usepackage{amsmath}
\usepackage[ruled,linesnumbered]{algorithm2e}
\usepackage{amsthm}

\usepackage{times}
\usepackage{amssymb}

\usepackage{caption}

\usepackage{booktabs}
\usepackage[sort&compress,numbers]{natbib}

\usepackage{url}
\usepackage[noblocks]{authblk}

\usepackage{xspace}
\newcommand*{\eg}{\emph{e.g.}\@\xspace}
\newcommand*{\ie}{\emph{i.e.}\@\xspace}
\newcommand*{\etal}{\emph{et al.}\@\xspace}

\makeatletter
\newcommand*{\etc}{%
    \@ifnextchar{.}%
        {etc}%
        {etc.\@\xspace}%
}
\makeatother

\title{Simple Primary Colour Editing for Consumer Product Images}

\author[1]{Han Gong}
\author[2]{Luwen Yu}
\author[2]{Stephen Westland}
\affil[1]{School of Computing Sciences, University of East Anglia, UK}
\affil[2]{School of Design, University of Leeds, UK}

\date{} 

\begin{document} 

\maketitle 

\thispagestyle{empty} 


\begin{abstract}
We present a simple primary colour editing method for consumer product images. We show that by using colour correction and colour blending, we can automate the pain-staking colour editing task and save time for consumer colour preference researchers. To improve the colour harmony between the primary colour and its complementary colours, our algorithm also tunes the other colours in the image. Preliminary experiment has shown some promising results compared with a state-of-the-art method and human editing.
\end{abstract}

%

\section{Introduction}
\label{sec:intro}
Research and data have been increasingly used to optimise the design process~\cite{yang2020predicting}. Previous research shows that product-colour appearance can affect consumers' purchase decisions while consumers' product-colour preferences vary from category to category    ~\cite{luo2019influence,yu2018role}. To understand consumer-product colour preference, the standard marketing images have been manually recoloured using software such as Adobe Photoshop~\cite{ps} or GIMP~\cite{gimp}. The recolouring process requires researchers (or designers) to manually adjust colours by picking colours and adopting non-binary per-layer masking. However, visible artefacts and incompatible background colours often remain even after very careful editing.
We conclude four main requirements from colour preference researchers:\\
\textbf{Minimum machine processing time}. Users would not prefer a slow processing speed as usually colour modifications are applied to multiple products for at least dozens of target colours.\\
\textbf{Minimum user manipulation time}. A high demand of user interaction time would be undesirable. Methods that require multiple user strokes or manual selection of multiple colours would not be ideal. We need a method that only requires a single primary colour specification and takes care of the rest of colour processing automatically.\\
\textbf{Artefact resiliency}. Artefacts, such as JPEG blocks and unnatural edges, are usually introduced after re-colouring. It is expected to preserve all image details except for the primary colour modification.\\
\textbf{Colour harmony preservation}. The chosen colour for change may not fit the product's existing complementary colours. In some cases, tuning of complementary colours is desirable.\\

\begin{figure*}[htb!]
\begin{center}
  \includegraphics[width=\linewidth]{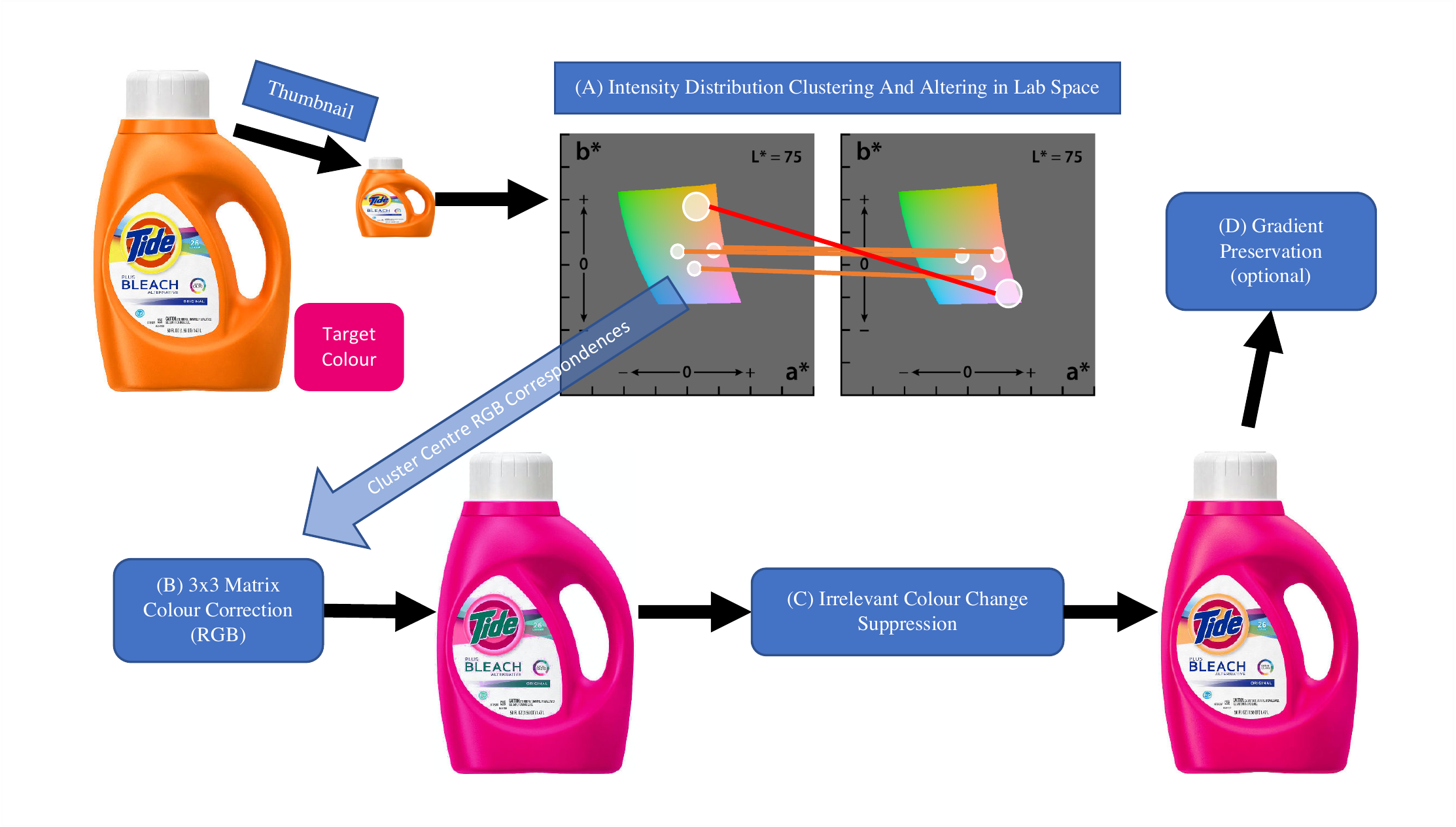}
\end{center}
\caption{Primary colour editing pipeline. Given an input image, our method amends the product's primary colour according to a target colour in 3-4 steps: A) Colour intensities clusters in CIE L*a*b* colour space~\cite{cie} are computed (left: original distributions; right: primary colour altered distributions); B) Cluster RGB correspondences are used for estimating a colour correction matrix.  Note that the L* channel intensities are also used for clustering but are not illustrated on the exemplar graphs; C) An alpha-blending process is applied to remove colour changes, which are less relevant to the primary colour change, from the colour corrected image; D) Some residual colour artefacts can be optionally removed using a gradient preservation method 'regrain'~\cite{Pitie2} (see the latter section for visualisation). The a*b* chromaticity gamut images are taken from Wikipedia~\cite{wiki_lab}.}
\label{fig:ppl}
\end{figure*}

Our proposed method in this paper addresses these requirements by providing an alternative design tool which is fully automatic. Existing studies suggested that colour manipulations offer the potential for software to generate recoloured images (target colour images). More promising applications of automatic colour manipulations will lead the trend of generative design systems in colour and design~\cite{sbai2018design,yang2020predicting}. 
There have been a number of methods for colour manipulations such as colour transfer~\cite{Pitie2,ReinhardTransfer,Nguyen}, colour hint propagation~\cite{farbman2010diffusion,chen2014sparse,chen2012manifold,an2008appprop,levin2004colorization}, or palette editing~\cite{nguyen2017group,chang2015palette,zhang2017palette}. However, none of the previous methods is directly applicable to the primary colour editing problem. Rapid digital workflows in practice would also require automatic methods for evaluating and/or comparing colours and designs~\cite{anderson2018design}.

In this paper, we propose a simple method which automates primary colour editing at an optimised consumption of user and machine processing time and it preserves colour harmony to some extent. Our method is based on the assumption that simulating colour change as a 2-D colour homography~\cite{finlayson2019color} (\ie as a change of light) usually avoids image processing  artefacts~\cite{gong2016recoding,finlayson2019color} such as JPEG blocks, sharp edges, and colour combination conflicts. Our colour editing pipeline is depicted in Figure~\ref{fig:ppl} where the colour editing task is reformulated as a 2-D colour homography colour correction problem. Additionally, we may apply a gradient preservation step to remove some residual artefacts. Compared with the previous recolouring methods, ours requires minimum user input and its design is relatively simple.

\section{Related Work}
\label{sec:bg}

Our work is relevant to the colour editing methods in three categories: A) colour transfer; B) colour hint propagation; C) palette-based colour editing.

\subsection{Colour transfer}
Colour transfer is an image editing process which adjusts the colours of a picture to match a target picture's colour theme. This research was started by Reinhard \etal~\cite{ReinhardTransfer} and followed up by the others~\cite{Nguyen,Pitie2,MKL_ct} recently. Most of these methods align the colour distributions in different colour spaces, which usually involve statistics alignment~\cite{ReinhardTransfer,Nguyen,MKL_ct} or iterative distribution shifting~\cite{Pitie2}.

\subsection{Colour hint propagation}
Some methods require user hints, \eg strokes, to guide recolouring of object surfaces. This direction of research was started by Levin \etal~\cite{levin2004colorization} where they colourise grey-scale images based on user colour stroke and solves for a large and sparse system of linear equations. Their key assumption is that the colours of neighbouring pixels with similar luminance should have similar chromaticities. More recent methods~\cite{an2008appprop,chen2012manifold,farbman2010diffusion} make use of  masks, either soft or hard, to assist re-colourisation. Their colour modification model is based on a diagonal colour correction matrix used for white balance, \eg \cite{CM} with limits on the range of applicable colour changes. Some others, \eg ~\cite{chen2014sparse}, have used sparse coding/learning that the sparse set of colour samples provide an intrinsic basis for an input image and the coding coefficients capture the linear relationship between all pixels and the samples. This branch of methods require heavy user inputs and therefore not immediately useful for our problem.
\subsection{Palette-based colour editing}
Some methods adopt colour intensity clustering, \eg k-means++ algorithm~\cite{kmeans++}, to initially generate a colour palette of the input image. After palette adjustments, different approaches were applied for manipulating colour changes. Zhang \etal~\cite{zhang2017palette} decompose the colours of the image into a linear combination of basis colours before reconstructing a new image using the linear coding coefficients. Chang \etal~\cite{chang2015palette} adopt a monotonic luminance mapping and radial basis functions (RBFs) for interpolating/mapping chromaticities. This branch of methods are most close to our solution however none of them is optimised for the particular task of rapid primary colour editing for consumer product images.

\subsection{Colour homography}
Our solution is based on the colour homography colour change model. The colour homography theorem~\cite{finlayson2019color,gong2016recoding,CIC2016,PICS2016} presents that chromaticities across a change in capture conditions (light color, shading and imaging device) are a homography apart. Suppose that we map an RGB $\underline{\rho}$ to a corresponding RGI (red-green-intensity) \underline{c} using a $3\times 3$ full-rank matrix $C$:
\begin{equation}
\begin{array}{c}
\underline{\rho}^\intercal C=\underline{c}^\intercal\\
\;\\
\left [
\begin{array}{c}
R\\
G\\
B
\end{array}
\right ]^\intercal
\left [
\begin{array}{ccc}
1 & 0 & 1\\
0 & 1 & 1\\
0 & 0 & 1
\end{array}
\right ]
=
\left [ 
\begin{array}{c}
R\\
G\\
R+G+B
\end{array}
\right ]^\intercal

\end{array}
\label{eq:chromaticity_conversion}
\end{equation}
The $r$ and $g$ chromaticity coordinates are written as ${r=R/(R+G+B)} \;,\;{g=G/(R+G+B)}$. We treat the right-hand-side of Equation~\ref{eq:chromaticity_conversion} as a homogeneous coordinate and we have $
\underline{c}\propto \left [
\begin{array}{ccc} r&g&1
\end{array}
\right]^\intercal
$.
When the shading is fixed, it is known that across a change in illumination or a change in device, the corresponding RGBs are approximately related by a $3 \times 3$ linear transform M that $\underline{\rho}^\intercal M = \underline{\rho}'^\intercal$ where $\underline{\rho}'$ is the corresponding RGBs under a second light or captured by a different camera~\cite{MARIMONT.WANDELL,MALONEY86B}. We have $H=C^{-1}MC$  which maps colours in RGI form between illuminants. Due to different shading, the RGI triple under a second light is written as $\underline{c}'^\intercal=\alpha\underline{c}^\intercal H$, where $\alpha$ denotes the unknown scaling. Without loss of generality we regard \underline{c} as a homogeneous coordinate \ie assume its third component is 1. Then, $[r'\;g']^\intercal=H([r\;g]^\intercal)$ (rg chromaticity coordinates are a homography $H()$ apart). In this paper, we will model the major colour change initially as a colour homography change but without considering the individual scale differences between each RGB correspondences, \ie a $3\times 3$ linear transform of colour change is applied.

\section{Simple Primary Colour Editing}
\label{sec:main}

Our algorithm starts with the simple observation that a simple 2-D colour homography model allows for a wider range of colour changes (as opposed to a diagonal colour correction matrix) and usually produces fewer colour combination conflicts~\cite{gong2016recoding,finlayson2019color}. In Figure~\ref{fig:ppl}, we overview the colour processing pipeline which consists of three major steps and one optional step: A) Clustering: The CIE L*a*b*~\cite{cie} intensities of an input RGB image are clustered using MeanShift~\cite{meanshift}. The primary colour cluster is altered to match the target colour (see the red line) that the cluster centres form the before-and-after sparse colour intensities correspondences; B) Colour correction: The L*a*b* colour correspondences are converted to RGB space before being used to estimate a 2-D colour homography matrix (without considering scale differences); C) Irrelevant colour change suppression: a soft alpha-blending mask is computed to suppress aggressive colour changes irrelevant to the primary colour change; D) Gradient preservation (optional): a gradient preservation step can be applied to remove more residual artefacts. We also note that the computational cost can be reduced by using down-sampled thumbnail images for model parameter estimation. We provide the algorithm details in the following sub-sections.

\subsection{Intensity clustering and altering}
To estimate a reliable colour change model, the first step is to extract the predominant colours which best capture the input image's colour theme. We adopt MeanShift~\cite{meanshift} clustering to extract at most 5 predominant colours (\ie cluster centres) from the input image. The intention of not collecting too many colours is to avoid noise and reduce computational cost. The cluster number of 5 is only an empirical value, \eg 6 also works. Clearly, a fixed set of MeanShift parameters never guarantee a maximum number of 5 colour clusters. We thus propose a simple adaptive MeanShift clustering procedure which gradually increases the initial small kernel bandwidth value as shown in Algorithm~\ref{alg:meanshift}
\begin{algorithm}
\SetAlgoLined
    $w=0.1$, $\beta=1.5$\;
    \Repeat{$n>5$}{
    $C=\mathbf{MeanShift}(w,A_{lab})$\;
    $n=\mathbf{len}(C)$\;
    $w=\beta w$\;
    }
\caption{Adaptive MeanShift clustering}
\label{alg:meanshift}
\end{algorithm}
where $\mathbf{MeanShift}$ is the MeanShift function with a flat kernel and bandwidth $w$, $C$ is a $n \times 3$ matrix of cluster centres (each row is a L*a*b* intensity vector), $\mathbf{len}$ counts the number of cluster centres $n$, $\beta$ is a factor controlling the kernel width growth rate in each iteration.

Given the obtained predominant colours, we construct the sparse colour correspondences to be supplied for colour change model estimation. Since we aim to only change the one primary colour if possible, the remaining of target predominant colours are kept the same as the original predominant colours except that the only primary colour is modified as the target primary colour. Through this, we construct a target predominant colour set denoted as $D$ (see also Figure~\ref{fig:ppl} (A) for illustration).

\subsection{Colour Homography colour change}
Given the source and target colour sets $C$ and $D$, we make use of a simple 2-D colour homography matrix to achieve primary colour change while minimising colour artefacts. A full colour homography change is an optimised chromaticity mapping in RGB space. However, since the brightness of colour matters in this application, we omit the shading factor $\alpha$ and only estimate a $3 \times 3$ linear matrix transform (which is still a homography matrix) using weighted least-squares as the follows: 
\begin{equation}
M =  \left(C^\intercal W C + kI_{3 \times 3}\right)^{-1} C^\intercal W D
\label{eq:lsq}
\end{equation}
where $k=10^{-3}$ is a regularisation term, $W$ is a diagonal matrix whose diagonal elements are the associated normalised weights of all the predominant colours (\ie cluster centre sizes), $I_{3 \times 3}$ is a $3 \times 3$ identity matrix. Denoting the 'flatten' RGB intensities of the input image as a $N \times 3$ matrix $A$ ($N$ is the number of pixels), we can compute its primary-colour-changed RGB intensities as $B=AM$. An intermediate processed example can be found in Figure~\ref{fig:ppl} (B).

\subsection{Irrelevant colour change suppression}
\begin{figure}[htb]
\begin{center}
  \includegraphics[width=\linewidth]{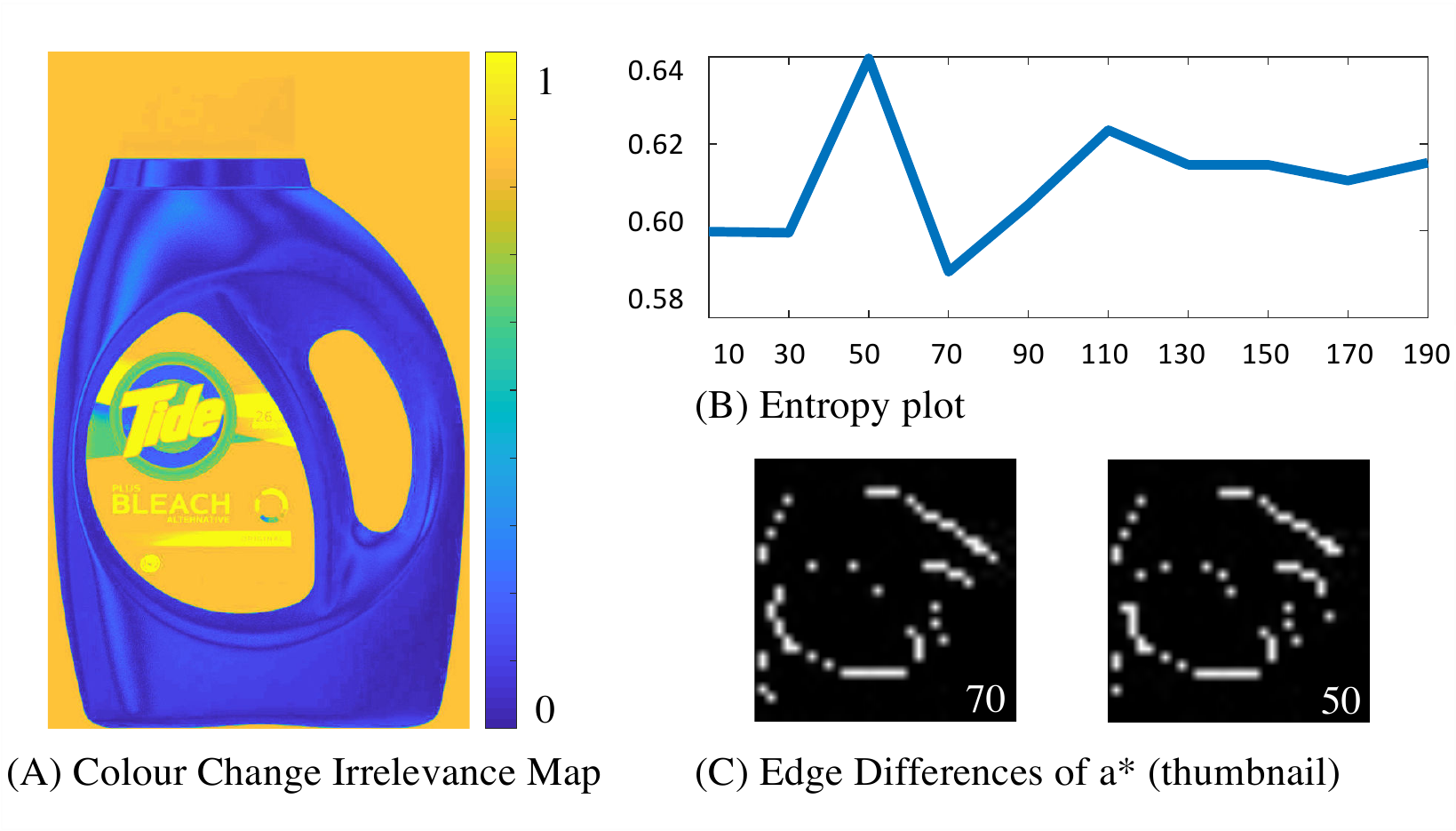}
\end{center}
\caption{Visualisations of the alpha blending. A) Visualised alpha blending mask used in the irrelevant colour change suppression step. B) The associated entropy plot (horizontal ticks: ${\Delta E}_{max}$). C) Visualisation of binary edge difference in channel a* when ${\Delta E}_{max}=70$ or ${\Delta E}_{max}=50$ (more different). The bottom-right number indicates the value of ${\Delta E}_{max}$. See also Figure~\ref{fig:ppl} (A) for the input image and the target primary colour.}
\label{fig:blending}
\end{figure}

Some of the colour changes after the $3 \times 3$ linear transform may look aggressive, \eg the pink ring of the 'Tide' logo in Figure~\ref{fig:ppl} (B). We adopt an alpha-blending procedure to address this as the follows:
\begin{equation}
B' = (1-\textbf{diag}(\underline{d})) B +  \textbf{diag}(\underline{d}) A
\label{eq:blending}
\end{equation}
where $B'$ is the modified RGB colour output, $\underline{d}$ is an N-vector denoting per-pixel scaling factors (in the range of $[0,1]$) and $\textbf{diag}()$ places an N-vector along the diagonal of an $N \times N$ diagonal matrix.
Our intuition is to smoothly reduce the impact of the colour changes that are irrelevant to the primary colour and control this by $\underline{d}$. We measure the irrelevance by the a*b* chromaticity difference $\Delta E$ between each colour (row) in $B'$ and the target primary colour:
\begin{equation}
\Delta E = \sqrt{ {\Delta a*}^{2} + {\Delta b*}^{2}}
\label{eq:ab_diff}
\end{equation}
where ${\Delta a*}$ and ${\Delta b*}$ are the errors in a*b* channels. A higher $\Delta E$ indicates a higher degree of irrelevance but this value can be sometimes too big. Thus, we further cap and normalise $\Delta E$ as $\Delta E'$:
\begin{equation}
\Delta E' = \left\{
\begin{array}{rcl}
1 &      & \Delta E > {\Delta E}_{max}\\
\Delta E/{\Delta E}_{max}  &      & \text{Otherwise}\\
\end{array} \right.
\label{eq:ab_diff_norm}
\end{equation}
where ${\Delta E}_{max}$ is an upper threshold value. The individual $\Delta E'$ is assigned as the corresponding element of $\underline{d}$. The processing result can be sensitive to ${\Delta E}_{max}$ and thus ${\Delta E}_{max}$ must be carefully chosen. An exemplar visualisation of $\underline{d}$ in its image grid form is shown in Figure~\ref{fig:blending} (A). Aiming at obtaining a blending result which preserves the edge details of the original image, we look for the optimum ${\Delta E}_{max}$ which minimises Equation~\ref{eq:ab_diff_best}. 
\begin{equation}
    \min_{{\Delta E}_{max}}\Sigma_{c \in \{a*, b*\}} \textbf{entropy}(| \textbf{edge}(I_{B',c}) - \textbf{edge}(I_{A,c}) |)
\label{eq:ab_diff_best}
\end{equation}
where $||$ is the operator to output per element absolute value of a matrix, $c$ indicates an intensity channel of a* or b*, $I_{B',*}$ and $I_{A,*}$ indicate the grid images of the 'unflatten' intensity matrix $B'$ and $A$ respectively, $\textbf{edge}$ is a binary edge detector using the Sobel approximation~\cite{sobel33x3} to the derivative (without edge-thinning), $\textbf{entropy}$ is a function which measures the amount of information -- entropy~\cite{entropy} -- as defined in Equation~\ref{eq:entropy}.
\begin{equation}
\textbf{entropy}(\underline{p}) = -\sum_i {p_i \log {p_i}}
\label{eq:entropy}
\end{equation}
where $\underline{p}$ is a normalised input vector (summed up to 1) which, in our case, is a 'flattened' error-of-edge image (\eg Figure~\ref{fig:blending} (C)), $i$ is an element index.
When the entropy of the error of two edge images is low, it indicates a higher similarity of edge features between two intensity images.
However, we do not have a closed-form solution for its global minima. In practice, a suitable local minima in a reasonable range usually serves the purpose. We thus propose a brutal search for a local minima solution of ${\Delta E}_{max}$ in the range of $[10,210]$ with an interval precision of $20$. A visualised example of $\underline{d}$ and its plot of ${\Delta E}_{max}$ search are shown in Figure~\ref{fig:blending}.

\subsection{Artefact cleansing}
\label{sec:main-art_removal}
\begin{figure}[htb]
\begin{center}
  \includegraphics[width=\linewidth]{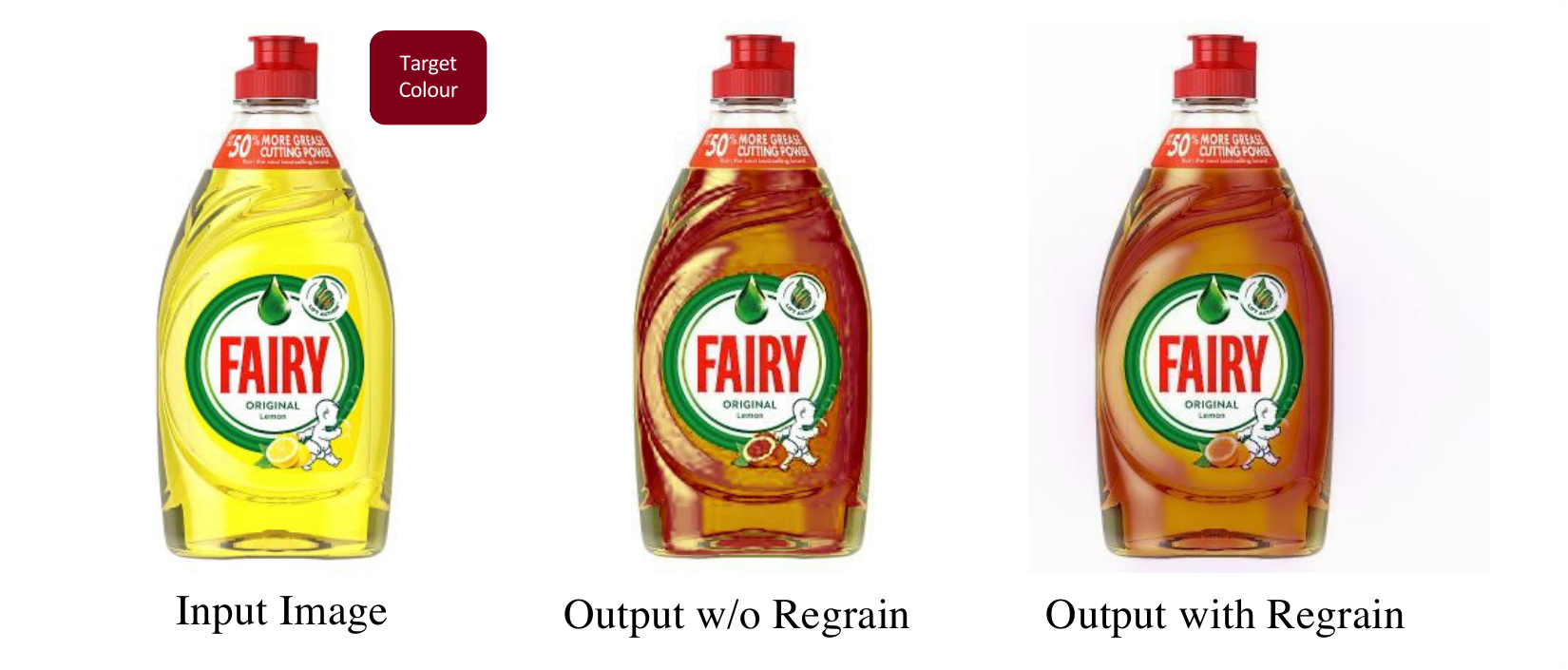}
\end{center}
\caption{Example of the 'regrain'~\cite{Pitie2} artefact cleansing.}
\label{fig:regrain}
\end{figure}

As the previous alpha blending step has attempted the minimisation of edge artefacts, mostly users can get an artefact-free output image. However, for some rare cases, we also adopt an optional artefact cleansing step called 'regrain' which was first proposed in \cite{Pitie2}. It provides strong gradient preservation but also has side effects which may cause minor undesired blurs along edges. Please refer to the cited paper for the algorithm details. Figure~\ref{fig:regrain} shows an example where this optional step improves the result by removing some JPEG block artefacts.
\subsection{Acceleration}
Our colour manipulation pipeline requires the solution of 10 key model parameters, namely $M$ and ${\Delta E}_{max}$. Using full-resolution images is not necessary and we therefore adopt thumbnail images ($32 \times 32$) for solving for $M$ and ${\Delta E}_{max}$ and apply the estimated parameters to a full-resolution input image to get a full-resolution output.

\section{Evaluation}
\label{sec:exp}
In this section, we present the result comparison and some useful discussions about our method's practical use-cases.

\subsection{Results}
\begin{figure*}
\begin{center}
  \includegraphics[width=\linewidth]{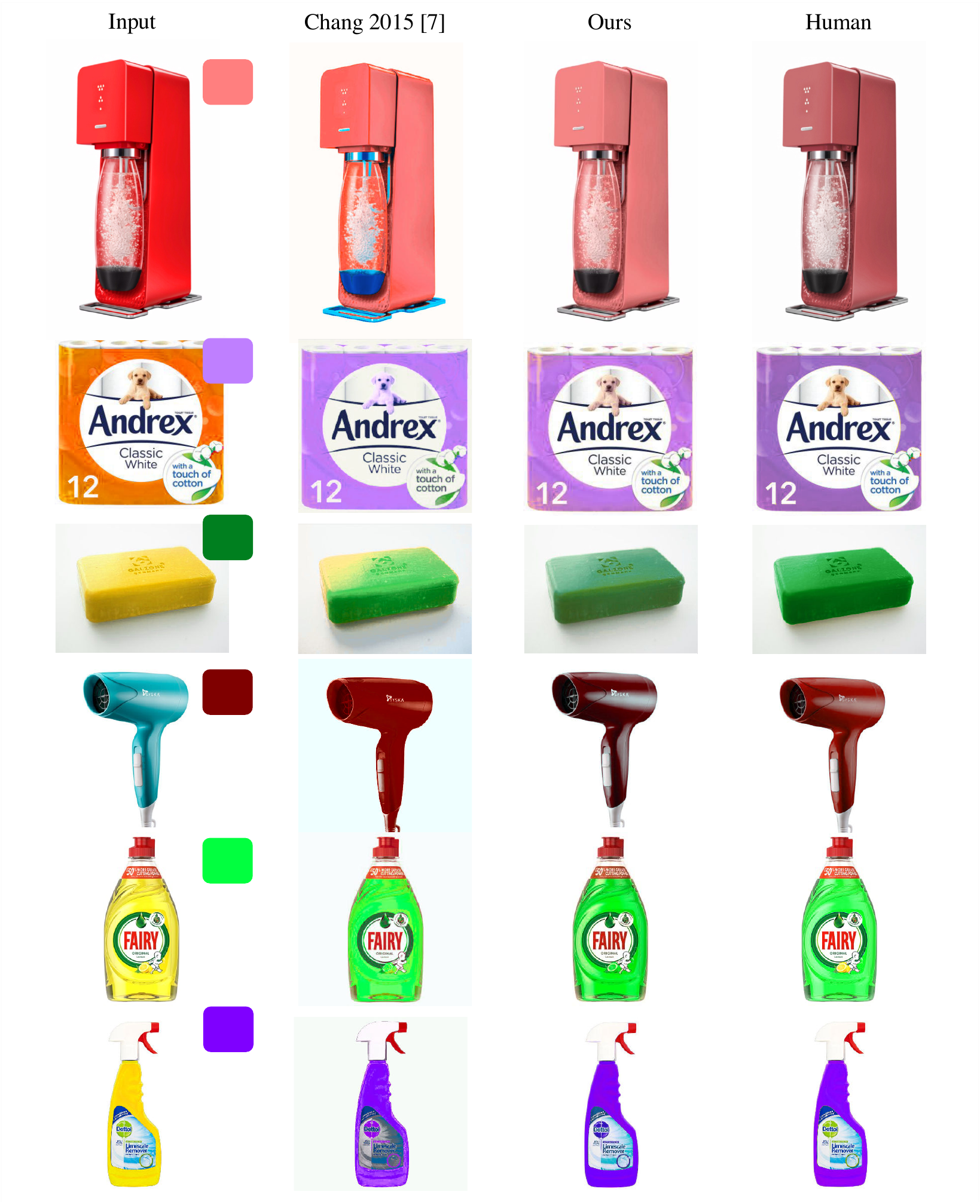}
\end{center}
\caption{Result visual comparison. The target colours are shown at the top right of the input images. The label 'Human' refers to the column of results manually generated by a professional colour preference researcher.}
\label{fig:res}
\end{figure*}

\begin{figure}[htb!]
\begin{center}
  \includegraphics[width=\linewidth]{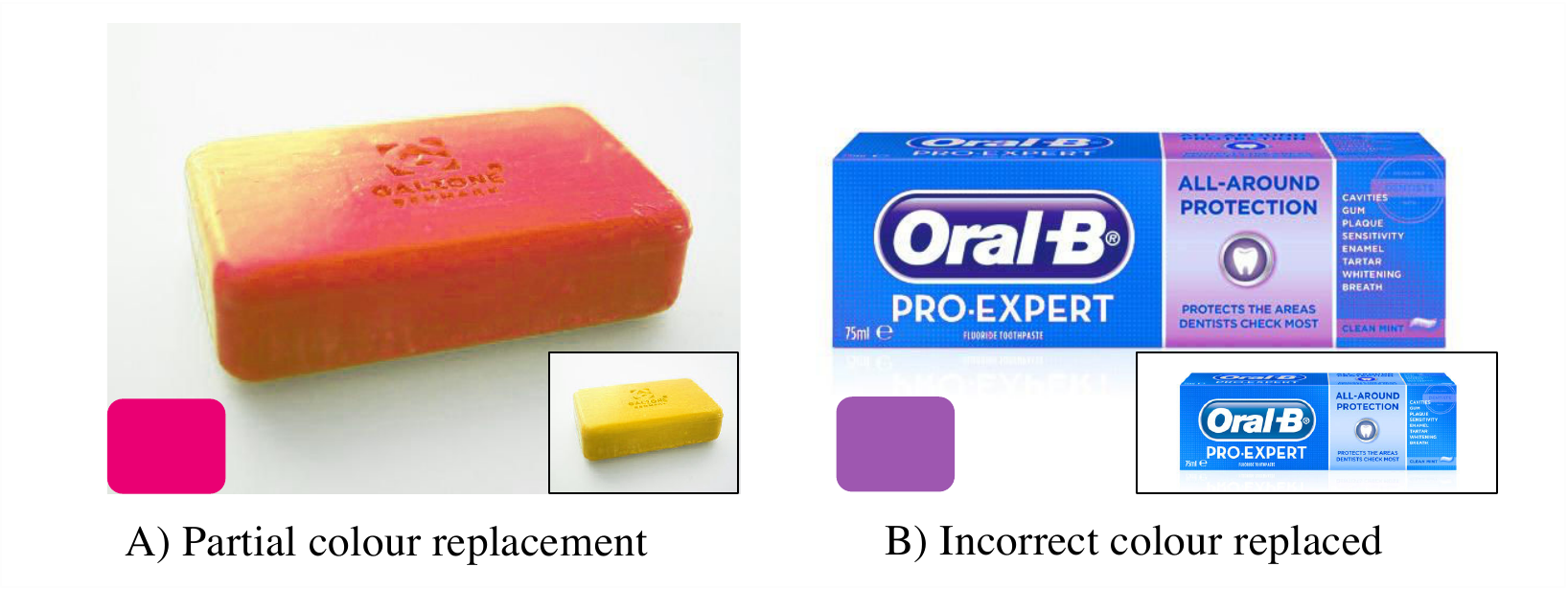}
\end{center}
\caption{Failure cases. The input images are shown at the bottom right and the target primary colours are shown at the bottom left. A) Only a part of the primary colour was replaced. B) An incorrect  primary colour was picked and replaced.}
\label{fig:failure}
\end{figure}

We compare our method with a state-of-the-art palette-based re-colouring method~\cite{chang2015palette} and the manually edited results produced by a professional colour preference researcher. Figure~\ref{fig:res} shows some visual result comparisons. We found that our outputs are mostly comparable to the manually edited results which take 2-5 minutes' labour time per image. Most of the human labour time is spent on masking the image (for primary colour pixels). Once the mask is completed, the remaining recolouring time takes about 1 minute. All the results in Figure~\ref{fig:res} have been produced without the 'regrain' step enabled.
Our method also has some failure cases as shown in Figure~\ref{fig:failure}. These failures were caused by the initial step of colour clustering. When the input image only has one colour, the MeanShift clustering algorithm can mishandle the primary colour extraction. Lowering the maximum cluster number (\ie 5 in Algorithm~\ref{alg:meanshift}) can resolve this issue. That said, we could provide this as an optional parameter for users.

Our method provides practical editing efficiency without user interventions. Using the thumbnail acceleration trick, our unoptimised MATLAB implementation (without the regrain step) takes about 1s to process an 1.2 mega-pixel image on a MacBook Pro 2015 laptop (2.5 GHz Quad-Core Intel Core i7 CPU).

\subsection{Discussions}
'Work to forecast' has suggested the use of colour to forecast consumer demand or resource-saving levels~\cite{schonberger2018reconstituting}. Colour has also been suggested as one of the most powerful visual elements in packaging. Thus, choosing an appropriate colour for the design of packaging or product can significantly affect consumer decision-making~\cite{kauppinen2014strategic}. This work could be applied as a product-colour predictor for studying product-packaging-colour in consumer purchase behaviour. Or, as an image generation tool, it helps designers/researchers preview the multi-colour options of a product image.

We also acknowledge that more rigorous user experiments in controlled lighting/display conditions could still be possibly carried out after the UK Covid-19 lockdown~\cite{horton2020offline}. We, therefore, commit to providing our source code to the research community in the hope that its evaluation and more potential use-cases can be further driven by the other cross-discipline communities.

\section{Conclusion}
\label{sec:con}
In this paper, we present a simple product re-colouring method for assisting consumer colour preference research. We show that by using a colour manipulation pipeline, we can automate this primary colour editing task for consumer colour preference researchers. The complementary colours in the product image will also be adjusted to potentially make the primary colour fits better. Future work is required to explore more of its use-cases and strengthen its artefact resiliency.

\section{Acknowledgment}
We also thank Dr Qianqian Pan from the University of Leeds for her useful discussions.

{\small
\bibliographystyle{plain}
\bibliography{biblio.bib}

\begin{thebibliography}{10}

\bibitem{gimp}
\emph{GNU Image Manipulation Program (GIMP)}, 2020.
\newblock \url{https://www.gimp.org} [Accessed: 06/04/2020].

\bibitem{wiki_lab}
\emph{L*a*b* gamut}, 2020.
\newblock \url{https://en.wikipedia.org/wiki/CIELAB_color_space} [Accessed:
  06/04/2020].

\bibitem{ps}
\emph{PhotoShop}, 2020.
\newblock \url{https://www.adobe.com/uk/products/photoshop.html} [Accessed:
  06/04/2020].

\bibitem{an2008appprop}
Xiaobo An and Fabio Pellacini.
\newblock Appprop: all-pairs appearance-space edit propagation.
\newblock In {\em ACM Transactions on Graphics (TOG)}, pages 1--9. 2008.

\bibitem{anderson2018design}
David Anderson, K~Blake Perez, Zack Xuereb, Kevin Otto, and Kris Wood.
\newblock Design processes of design automation practitioners.
\newblock In {\em International Design Engineering Technical Conferences and
  Computers and Information in Engineering Conference}. American Society of
  Mechanical Engineers Digital Collection, 2018.

\bibitem{kmeans++}
David Arthur and Sergei Vassilvitskii.
\newblock k-means++: The advantages of careful seeding.
\newblock Technical report, Stanford, 2006.

\bibitem{chang2015palette}
Huiwen Chang, Ohad Fried, Yiming Liu, Stephen DiVerdi, and Adam Finkelstein.
\newblock Palette-based photo recoloring.
\newblock {\em ACM Transactions on Graphics}, 34(4):139--1, 2015.

\bibitem{chen2014sparse}
Xiaowu Chen, Dongqing Zou, Jianwei Li, Xiaochun Cao, Qinping Zhao, and Hao
  Zhang.
\newblock Sparse dictionary learning for edit propagation of high-resolution
  images.
\newblock In {\em Proceedings of the IEEE Conference on Computer Vision and
  Pattern Recognition}, pages 2854--2861, 2014.

\bibitem{chen2012manifold}
Xiaowu Chen, Dongqing Zou, Qinping Zhao, and Ping Tan.
\newblock Manifold preserving edit propagation.
\newblock {\em ACM Transactions on Graphics (TOG)}, 31(6):1--7, 2012.

\bibitem{farbman2010diffusion}
Zeev Farbman, Raanan Fattal, and Dani Lischinski.
\newblock Diffusion maps for edge-aware image editing.
\newblock {\em ACM Transactions on Graphics}, 29(6):1--10, 2010.

\bibitem{PICS2016}
Graham~D. Finalyson, Han Gong, and Robert~B. Fisher.
\newblock Color homography.
\newblock In {\em Progress in Colour Studies}. John Benjamins Publishing
  Company, 2016.

\bibitem{CIC2016}
Graham~D. Finalyson, Han Gong, and Robert~B. Fisher.
\newblock Color homography color correction.
\newblock In {\em Color Imaging Conference}. Society for Imaging Science and
  Technology, 2016.

\bibitem{finlayson2019color}
Graham Finlayson, Han Gong, and Robert~B Fisher.
\newblock Color homography: theory and applications.
\newblock {\em IEEE transactions on pattern analysis and machine intelligence},
  41(1):20--33, 2019.

\bibitem{meanshift}
Keinosuke Fukunaga and Larry Hostetler.
\newblock The estimation of the gradient of a density function, with
  applications in pattern recognition.
\newblock {\em IEEE Transactions on information theory}, 21(1):32--40, 1975.

\bibitem{CM}
Han Gong.
\newblock Convolutional mean: A simple convolutional neural network for
  illuminant estimation.
\newblock In {\em British Machine Vision Conference}. BMVA, 2019.

\bibitem{gong2016recoding}
Han Gong, Graham~D. Finlayson, and Robert~B. Fisher.
\newblock Recoding color transfer as a color homography.
\newblock In {\em British Machine Vision Conference}. BMVA, 2016.

\bibitem{horton2020offline}
Richard Horton.
\newblock Offline: Covid-19 and the nhs—“a national scandal”.
\newblock {\em The Lancet}, 395(10229):1022, 2020.

\bibitem{kauppinen2014strategic}
Hannele Kauppinen-R{\"a}is{\"a}nen.
\newblock Strategic use of colour in brand packaging.
\newblock {\em Packaging Technology and Science}, 27(8):663--676, 2014.

\bibitem{levin2004colorization}
Anat Levin, Dani Lischinski, and Yair Weiss.
\newblock Colorization using optimization.
\newblock In {\em ACM Transactions on Graphics}, pages 689--694. 2004.

\bibitem{luo2019influence}
Dan Luo, Luwen Yu, Stephen Westland, and Nik Mahon.
\newblock The influence of colour and image on consumer purchase intentions of
  convenience food.
\newblock {\em Journal of the International Colour Association}, 24:11--23,
  2019.

\bibitem{MALONEY86B}
{L.T.} Maloney.
\newblock {E}valuation of linear models of surface spectral reflectance with
  small numbers of parameters.
\newblock {\em Journal of the Optical Society of America A}, 3:1673--1683,
  1986.

\bibitem{MARIMONT.WANDELL}
{D.H.} Marimont and {B.A.} Wandell.
\newblock {L}inear models of surface and illuminant spectra.
\newblock {\em Journal of the Optical Society of America A}, 9(11):1905--1913,
  92.

\bibitem{Nguyen}
R.~M.~H. Nguyen, S.~J. Kim, and M.~S. Brown.
\newblock Illuminant aware gamut-based color transfer.
\newblock {\em Computer Graphics Forum}, 33(7):319--328, October 2014.

\bibitem{nguyen2017group}
Rang~MH Nguyen, Brian Price, Scott Cohen, and Michael~S Brown.
\newblock Group-theme recoloring for multi-image color consistency.
\newblock In {\em Computer Graphics Forum}, volume~36, pages 83--92. Wiley
  Online Library, 2017.

\bibitem{MKL_ct}
F~Piti{\'e} and A~Kokaram.
\newblock The linear monge-kantorovitch linear colour mapping for example-based
  colour transfer.
\newblock In {\em European Conference on Visual Media Production}. IET, 2007.

\bibitem{Pitie2}
Fran\c{c}ois Piti{\'e}, Anil~C. Kokaram, and Rozenn Dahyot.
\newblock Automated colour grading using colour distribution transfer.
\newblock {\em Computer Vision And Image Understanding}, 107(1-2):123--137,
  July 2007.

\bibitem{ReinhardTransfer}
Erik Reinhard, Michael Ashikhmin, Bruce Gooch, and Peter Shirley.
\newblock Color transfer between images.
\newblock {\em IEEE Computer Graphics And Applications}, 21(5):34--41,
  September 2001.

\bibitem{sbai2018design}
Othman Sbai, Mohamed Elhoseiny, Antoine Bordes, Yann LeCun, and Camille
  Couprie.
\newblock Design: Design inspiration from generative networks.
\newblock In {\em European Conference on Computer Vision}, pages 0--0, 2018.

\bibitem{cie}
Janos Schanda.
\newblock Cie colorimetry.
\newblock {\em Colorimetry: Understanding the CIE system}, pages 25--78, 2007.

\bibitem{schonberger2018reconstituting}
Richard~J Schonberger.
\newblock Reconstituting lean in healthcare: From waste elimination toward
  ‘queue-less’ patient-focused care.
\newblock {\em Business Horizons}, 61(1):13--22, 2018.

\bibitem{entropy}
Claude~E Shannon.
\newblock A mathematical theory of communication.
\newblock {\em Bell system technical journal}, 27(3):379--423, 1948.

\bibitem{sobel33x3}
I~Sobel and G~Feldman.
\newblock A 3x3 isotropic gradient operator for image processing, presented at
  a talk at the stanford artificial project.
\newblock {\em Pattern Classification and Scene Analysis}, pages 271--272, 3.

\bibitem{yang2020predicting}
Jie Yang, Yun Chen, Stephen Westland, and Kaida Xiao.
\newblock Predicting visual similarity between colour palettes.
\newblock {\em Color Research \& Application}, 2020.

\bibitem{yu2018role}
Luwen Yu, Stephen Westland, Zhenhong Li, Qianqian Pan, Meong~Jin Shin, and
  Seahwa Won.
\newblock The role of individual colour preferences in consumer purchase
  decisions.
\newblock {\em Color Research \& Application}, 43(2):258--267, 2018.

\bibitem{zhang2017palette}
Qing Zhang, Chunxia Xiao, Hanqiu Sun, and Feng Tang.
\newblock Palette-based image recoloring using color decomposition
  optimization.
\newblock {\em IEEE Transactions on Image Processing}, 26(4):1952--1964, 2017.

\end{thebibliography}
}

\end{document}